%% file: paper.tex
\documentclass[letterpaper, 10 pt, conference]{ieeeconf}  

\IEEEoverridecommandlockouts                              

\usepackage{algorithm}
\usepackage{algpseudocode}
\usepackage{listings}
\usepackage{fancyhdr,graphicx,amsmath,amssymb}
\usepackage{xcolor}
\usepackage{amsfonts}
\usepackage{siunitx}
\usepackage{multirow}
\usepackage{bm}
\usepackage{mathrsfs}
\usepackage{amsmath}

\definecolor{ieeeblue}{RGB}{0,98,155}

\newcommand\Vmax{V_{max}}
\newcommand\Dmin{D_{min}}
\newcommand\Lmin{L_{i,min}}
\newcommand\Lmax{L_{i,max}}
\newcommand\Limin{L_{i,min}}
\newcommand\Limax{L_{i,max}}
\newcommand\Ljmin{L_{j,min}}

\newcommand\WP{\bm{\omega}_p}
\newcommand\W{\bm{\omega}}
\newcommand\Wip{\bm{\omega}_{i/p}}
\newcommand\wip{\omega_{i/p}}
\newcommand\Wio{\bm{\omega}_{i/o}}
\newcommand\wio{\omega_{i/o}}
\newcommand\Wop{\bm{\omega}_{o/p}}
\newcommand\w{\omega}
\newcommand\wP{\omega_p}

\newcommand\V{\mathbf{v}}
\newcommand\vBi{v_{B_i}}
\newcommand\VP{\mathbf{v}_p}
\newcommand\vP{v_p}
\newcommand\Bi{\mathbf{b}_i}
\newcommand\bi{b_i}
\newcommand\norm[1]{\| #1 \|_2}
\newcommand\Fi{\mathscr{F}_i}

\newcommand\straightpath[0]{\textit{straight path}}
\newcommand\straightpaths[0]{\textit{straight paths}}
\newcommand\crossprod[1]{\left[#1\right]_{\times}}

\title{\LARGE \bf
Continuous Collision Detection for a Robotic Arm Mounted on a Cable-Driven Parallel Robot
}

\author{Diane Bury$^{1,2}$, Jean-Baptiste Izard$^{1}$ , Marc Gouttefarde$^{3}$, Florent Lamiraux$^{2}$
\thanks{$^{1}$Tecnalia France, Bat 6 CSU, 950 rue Saint-Priest, 34090 Montpellier, France
        {\tt\small diane.bury@tecnalia.com},
        {\tt\small jeanbaptiste.izard@tecnalia.com}}%
\thanks{$^{2}$LAAS-CNRS, University of Toulouse, Toulouse, France
        {\tt\small florent.lamiraux@laas.fr}}%
\thanks{$^{3}$LIRMM, Universit\'e de Montpellier, CNRS, Montpellier, France
        {\tt\small marc.gouttefarde@lirmm.fr}}%
}

\begin{document}

\maketitle
\thispagestyle{empty}
\pagestyle{empty}

\begin{abstract}
A continuous collision checking method for a cable-driven parallel robot with an embarked robotic arm is proposed in this paper. The method aims at validating paths by checking for collisions between any pair of robot bodies (mobile platform, cables, and arm links). For a pair of bodies, an upper bound on their relative velocity and a lower bound on the distance between the bodies are computed and used to validate a portion of the path. These computations are done repeatedly until a collision is found or the path is validated. The method is integrated within the Humanoid Path Planner (HPP) software, tested with the cable-driven parallel robot CoGiRo, and compared to a discretized validation method.
\end{abstract}

\section{INTRODUCTION}
\input {intro}

\input {continuous-collision-checking}

\section {IMPLEMENTATION RESULTS IN HPP}\label{results}
\input {experimental-results}

\section{CONCLUSIONS}\label{conclusion}

This paper introduced a method which extends a continuous collision checking method for a multi-arm robot to a CDPR with a robotic arm mounted on its platform, to take into account collisions including the cables. The method has been integrated within the software HPP and tested in simulations on the robot CoGiRo. Results show that the method is effective and is able to find collisions which are undetected by a discretized method even with a small time step. Although our method is exact and can validate a path regarding collisions, it does not ensure the path is adapted to be performed on the physical robot. The algorithm could be improved using a cost-base planning method to maximize distance to obstacles. Future work will include applications on the physical robot CoGiRo to plan movements in a cluttered environment.

\begin{center}

\begin{table}[t!]
\resizebox{\columnwidth}{!}{%
  \begin{tabular}{|l|l|l|l||l|l|l|}
    \hline
	  $ \ $ computing &     
      \multicolumn{3}{c||}{Continuous} &
      \multicolumn{3}{c|}{Discretized, $\tau$ = 0.1s} \\
    & & & & & & \\[-1em]
    $ \ $ time (s) & min & mean & max & min & mean & max \\
    \hline
    & & & & & & \\[-1em]
    True positives & 1.8e-04 & 0.095 & 1.1 & 2.6e-04 & 9.7e-03 & 0.052 \\
    \hline
    & & & & & & \\[-1em]
    True negatives & 0.18 & 0.57 & 2.3 & 0.011 & 0.033 & 0.057 \\
    \hline
    & \multicolumn{1}{c}{} & \multicolumn{1}{c}{} & & \multicolumn{1}{c}{} & \multicolumn{1}{c}{} & \\[-1em]
    \textbf{All paths} & \multicolumn{1}{c}{} & \multicolumn{1}{c}{0.27} & \multicolumn{1}{c||}{} & \multicolumn{1}{c}{} & \multicolumn{1}{c}{0.018} & \multicolumn{1}{c|}{} \\
    \hline
  \end{tabular}%
}
  \vspace{-2mm}
\end{table}

\begin{table}[t!]
\resizebox{\columnwidth}{!}{%
  \begin{tabular}{|l|l|l|l||l|l|l|}
    \hline
	  $ \ $ computing &     
      \multicolumn{3}{c||}{Discretized, $\tau$ = 0.01s} &
      \multicolumn{3}{c|}{Discretized, $\tau$ = 0.001s} \\
    & & & & & & \\[-1em]
    $ \ $ time (s) & min & mean & max & min & mean & max \\
    \hline
    & & & & & & \\[-1em]
    True positives & 2.6e-04 & 0.095 & 0.50 & 2.6e-04 & 0.92 & 4.6 \\
    \hline
    & & & & & & \\[-1em]
    True negatives & 0.11 & 0.33 & 0.59 & 1.0 & 3.3 & 6.7 \\
    \hline
    & \multicolumn{1}{c}{} & \multicolumn{1}{c}{} & & \multicolumn{1}{c}{} & \multicolumn{1}{c}{} & \\[-1em]
    \textbf{All paths} & \multicolumn{1}{c}{} & \multicolumn{1}{c}{0.18} & \multicolumn{1}{c||}{} & \multicolumn{1}{c}{} & \multicolumn{1}{c}{1.8} & \multicolumn{1}{c|}{} \\
    \hline
  \end{tabular}%
}
\vspace{2mm}
  \caption{Computing times for 1000 random \straightpaths{}}
  \label{table:computing_times}
\vspace{-8mm}
\end{table}

\end{center}

\addtolength{\textheight}{-12cm}   





\bibliography {paper}
\bibliographystyle{IEEEtran}

\end{document}

%% file: intro.tex
\begin {figure}[t!]
  \centerline {
    \includegraphics[width=0.45\textwidth]{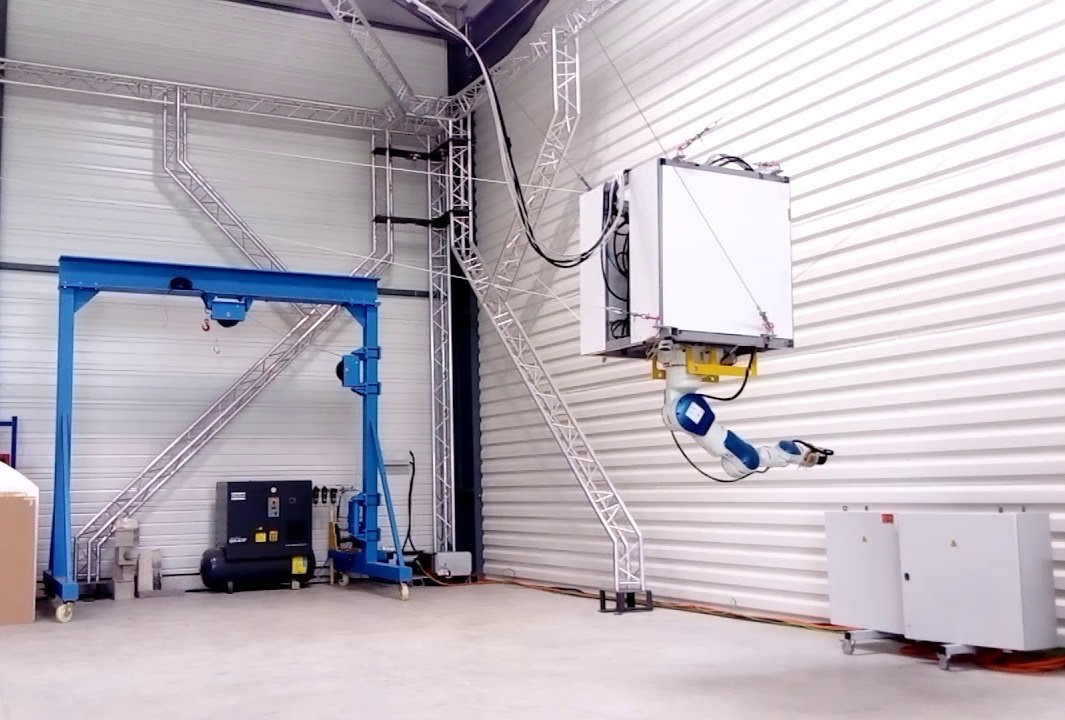}
  }
  \caption {Cable-driven parallel robot CoGiRo with a 7-DOF robotic arm.}
  \label {fig:robot}
  \vspace{-5mm}
\end {figure}

This paper deals with continuous collision checking for a cable-driven parallel robot (CDPR) with a robotic arm mounted on it. 
The method developped in this paper is applied to the robot CoGiRo, shown in Fig. \ref{fig:robot}, a redundantly-actuated cable-suspended CDPR developed by LIRMM and Tecnalia. The problem consists in determining exactly whether an input path between two configurations of the robot (consisting of the CDPR and the arm) is collision-free, or if there exists a configuration along the path for which one of the robot body or cable is in collision with another robot body, cable or static part of the environment. Although for academic instances of path planning problems, discretized collision checking is usually enough, exact or continuous collision checking is very important when dealing with real robots in industrial settings. In some applications like path length optimization for instance~\cite{CamLamLau2016}, it is important that if a path has been validated, any sub-path be computed as valid using the same validation algorithm. This is not the case with discretized collision checking since the samples tested for collision may not be the same.

The developed method is integrated within the existing open-source software Humanoid Path Planner (HPP) \cite{MirTonFerSepCamManLam2016} which includes sampling-based planning algorithms such as Probabilistic RoadMaps (PRM) and Rapidly-exploring Random Trees (RRT), and different optimization methods.

Several previous works on CDPRs dealt with the issues of cable-cable, cable-platform and cable-object collisions. The determination of the loci of cable collisions within a prescribed workspace is presented in \cite{MerletARK2004,Perreault_Cardou_JMR2010} in the case of a constant-orientation workspace and in \cite{Blan2015,BlaMer2014,NguyenCableCon2014} in the case of a 6D workspace. Fast heuristic approaches are proposed in \cite{NguyenCableCon2014} whereas certified calculations based on interval analysis are introduced in \cite{Blan2015,BlaMer2014}.
Besides, an approximate determination of the volume swept by a cable when the mobile platform of a CDPR moves within a prescribed workspace is discussed in \cite{PottCableSpanCableCon2017} and the geometric determination of cable-cylinder collision loci within the workspace of a CDPR is dealt with in \cite{MartinCableCon2017}. In the latter, the cylinder is a fixed object located inside the CDPR workspace. Moreover, in \cite{Fabritius2018}, the collision-free printing workspace is calculated for fully-constrained CDPRs intended to print large-dimension objects in a sequence of horizontal layers. While all these methods determine various types of cable collision loci within a prescribed workspace, other previous works address the issue of checking cable collisions along a prescribed CDPR mobile platform path \cite{Blan2015, MerletDaney_ICRA2006,Lahouar2009}, which is also the purpose of the present paper.
In \cite{Lahouar2009}, a classic discretized collision checking applied to CDPRs is presented. More advanced methods based on interval analysis, which can account for parameter uncertainties and round-off errors in numerical calculation, are introduced in \cite{Blan2015, MerletDaney_ICRA2006}.

The contribution of the present paper is a continuous cable collision checking method integrated within the open-source software HPP which can test collisions along a prescribed path of the mobile platform of a CDPR. Compared to \cite{Lahouar2009}, the proposed method allows continuous collision checking instead of discretized checking. The methods presented in \cite{Blan2015,MerletDaney_ICRA2006} to check cable collisions along a path can be classified as continuous checking method. Indeed, a proper use of interval analysis ensures that there does not exist any robot configuration along the path where a cable collision can occur, taking into account model parameter uncertainties and round-off errors. However, to the best of our knowledge, these methods have not yet been applied to a problem case similar to ours. In this paper, we propose a method that takes into acount not only collisions between the cables, but also between the cables and all other bodies of the robot or the environment. Our method is also easily extended to other types of continuous validation of a path, for example cable tension validation. Moreover, its implementation in HPP makes it easy to use with any new robot, by providing the necessary files modeling the new robot. The contribution of the present paper can thus be used with any CDPR.

Besides, Schwarzer \cite {SchSahLat2004} proposes a method to check collision of a multi-arm robot along a continuous path composed of linear interpolations in the joint space. The method is based on the computation of upper bounds on the relative velocities -- linear and angular velocities -- of each body in the reference frame of the other bodies. Then, given the distance -- or a lower bound on the distance -- between two bodies at a given parameter value, the method computes a time interval over which no collision can occur between the bodies. This method is already implemented in HPP.

Cables are not solid bodies because they deform over time: they are subject to elongation and sagging. Even when assimilating the cable to a perfect cylinder attached at one end to the mobile platform and at the other end to a fixed exit point on the structure base, the cable length varies depending on the pose of the platform. Collisions between the cables and the mobile platform are possible and must be checked, but since the cables are attached to the platform, the lower bound on the distance is zero. For those reasons, \cite {SchSahLat2004} cannot be directly applied to a CDPR.

The method proposed in the present paper is inspired from \cite {SchSahLat2004} and adapted to CDPR cable collisions. The method takes as input
\begin {itemize}
\item A set of \textit{collision elements} that are constituted by pairs of objects among robotic arm bodies, platform and cables of the CDPR, and static obstacles in the environment.
\item A path which is a linear interpolation in the joint configuration space, called in this paper a \straightpath{}. The joint space of the CDPR consists of the 7 values for the pose of the mobile platform (3 values for the translation and a quaternion for the rotation) and of the values of the joints of the robotic arm.
\end {itemize}

In Section \ref{continuous-collision-checking}, we present useful definitions and notations, and the overall algorithm used to validate a continuous path. Section \ref{boundcalculation} details the different methods used to calculate the velocity upper bound and the distance lower bound for the different types of pair of bodies. Implementation results in HPP are described in Section \ref{results} and Section \ref{conclusion} concludes the paper.

%% file: continuous-collision-checking.tex
\section {CONTINUOUS COLLISION CHECKING}\label{continuous-collision-checking}

\subsection{Definitions and notations}\label{sec:definitions}

\begin {figure}
  \centerline {
    \includegraphics[width=0.45\textwidth]{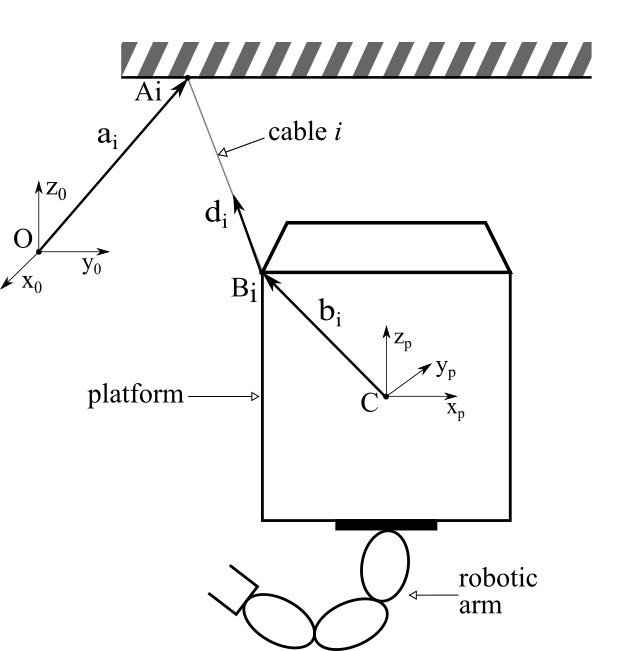}
  }
  \caption {Notations}
  \label{fig:notations}
\end {figure}

We consider a CDPR composed of a mobile platform suspended by cables attached to a fixed frame structure, as depicted in Fig. \ref{fig:notations}. An articulated robotic arm is mounted on the platform for the purpose of grabing and moving objects in the environment. Some simplifications are made to modelize the cables:
\begin{itemize}
\item The cables are considered to be cylinders.
\item Sagging of the cables is either neglected or taken into account in the cylinder radius.
\item Each cable exits the fixed structure at a fixed point $A_i$.
\item Each cable is attached to the mobile platform at a point $B_i$, fixed in the platform frame.
\end{itemize}

During motion planning, \straightpaths{} are computed and must be checked against collisions. We assume that along a \straightpath{}, the platform rotates or translates at a constant linear or angular velocity --- respectively $\VP$ and $\WP$ --- in the fixed reference frame. To validate a path in regard to collisions, every pair of bodies has to be checked for collisions. Those pairs are called collision elements, and can be of different types: a cable-cable collision, a cable-platform collision, a cable-arm collision, or robot-robot collision. In the software, collision element types are represented as derived classes of an abstract class, \texttt{\textsc{CollisionElement}}. The path to validate is set at the very start of the algorithm for all collision elements. This class has a method \texttt{\textsc{validateElement}}, which takes as input the time parameter value $t$ corresponding to the configuration to validate. The output is a Boolean value indicating whether or not the configuration at the time parameter $t$ is valid, in which case the method also returns a collision-free interval containing $t$, denoted as $interval$.
The size of $interval$ is computed using an upper bound on the velocity $\Vmax$ of one of the bodies of the pair relative to the other, and a lower bound on the distance between the two bodies $\Dmin$. The calculations for those bounds differ depending on the type of the collision element as detailed in section \ref{boundcalculation}.

\begin{algorithm}[t!]
  \caption{Validation of a \straightpath{} using the dichotomy method}\label{dichotomy}
  \begin{algorithmic}[1]
    \Function{ValidateStraightPath}{$path$}
    \State $t \gets 0$
    \State $validSubset \gets \emptyset$
    \State $valid \gets$ True
    \While{$valid$ is True and $validSubset\not=[0,T]$}
 		\State $success, validInterval \gets$ \Call{ValidateAllCollisionElements}{$t$}
 		\If{not $success$}
 			\State $valid \gets$ False
 		\Else
 			\State $validSubset \gets validInterval \cup validSubset$
 		\EndIf
 		\State $t \gets$ middle of first interval of $\overline{validSubset}$
    \EndWhile
    \If{not $valid$}
    	\State \Return first interval of $validSubset$
	\Else
		\State \Return $[0,T]$
    \EndIf
    \EndFunction
  \end{algorithmic}
\end{algorithm}

\subsection{Continuous validation algorithm}\label{algo}

\subsubsection{Validation of a \straightpath{}}

The input path to validate is a concatenation of linear interpolation paths that are called \straightpaths{}. Thus we need to successively validate each \straightpath{} using the function \texttt{\textsc{ValidateStraightPath}}, until a collision is found or until all \straightpaths{} have been validated.

Let us denote by $[0,T]$ the interval of definition of the \straightpath{} to validate. If $I$ is a subset of real numbers, we denote by $\overline{I}$ the complement of $I$ in $[0,T]$. Algorithm \ref{dichotomy} validates a path by creating a subset of validated intervals, $validSubset$. This subset is initialized with the empty set. The algorithm then loops until $validSubset$ is equal to $[0,T]$. $t$ is initialized to $0$. Function \texttt{\textsc{ValidateAllCollisionElements}} is called with $t$ and returns an interval containing $t$ valid for all the collision elements. $validSubset$ is augmented with this newly validated interval (the new $validSubset$ is the union of the previous $validSubset$ and the newly validated interval). Then $t$ receives the parameter value of the middle of the first non-tested interval. The algorithm then progresses until either $[0,T]$ is entirely validated, or a collision is found.

\subsubsection{Finding a valid interval for all collision elements around a given configuration}

As detailed in Algorithm \ref{validateallelements}, the function \texttt{\textsc{ValidateAllCollisionElements}} takes as input a valid parameter value in $[0,T]$ and returns a valid interval containing the input parameter value. The function loops over all the collision elements, stored in a list $collisionElemList$. A local variable $validInterval$ is initialized to $]-\infty;+\infty[$ and represents the interval currently validated. For each $collisionElem$, an interval valid for this collision element only is calculated with \texttt{\textsc{ValidateElement}} and the total valid interval $validInterval$ is trimmed by doing an intersection with this newly calculated interval.

\begin{algorithm*}[t!]
  \caption{Validation of a configuration by validating each interval element}\label{validateallelements}
  \begin{algorithmic}[1]
    \Function{ValidateAllCollisionElements}{$t$}
    	\State $validInterval \gets ] - \infty,  \infty [$
		\For{each collision element $collisionElem$ in $collisionElemList$}
			\State $(valid, newValidInterval, report) \gets$ \Call{ValidateElement}{$collisionElem$, $t$, $validInterval$}
			\If{not $valid$} \Comment{there is a collision for $config(t)$}
				\State \Return $($False$, validInterval, report)$
			\Else \Comment{there is no collision for $config(t)$}
				\State $validInterval \gets validInterval \cap newValidInterval$ 
			\EndIf
		\EndFor
    	\State \Return (True, $validInterval$) \Comment{$validInterval$ is continuously valid for every collision element}
    \EndFunction
  \end{algorithmic}
\end{algorithm*}


\subsubsection{Validating an interval for one collision element}

The function \texttt{\textsc{ValidateElement}} returns an interval around the given time parameter $t$ for a collision element. Each collision element stores the set of intervals currently validated for a path. This variable is reset each time the algorithm starts working on a new path. Before computing a valid interval around $t$, it is first checked if the interval to validate has already been validated.

If not, an upper bound $\Vmax$ of the velocity of one of the bodies relative to the other body of the pair is calculated, as well as a lower bound $\Dmin$ of the distance at parameter $t$ between the two bodies of the element (see next section for the calculation methods). If a collision is found during the calculation of $\Dmin$, \texttt{\textsc{ValidateElement}} returns $false$ and reports the collision. If no collision is found, the value of $\Dmin$ is a strictly positive real number. We define the output valid interval $newValidInterval$ in (\ref{eq:halflength}).


\vspace{-4mm}
\begin{equation}\label{eq:halflength}
newValidInterval =  [ t - \frac{\Dmin}{\Vmax} , t + \frac{\Dmin}{\Vmax}]
\end{equation}


\section{CALCULATION OF THE VELOCITY AND DISTANCE BOUNDS}
\label{boundcalculation}

\subsection{Collision between two bodies of the robot}

As previously stated, for a pair of non-cable bodies of the robot (mobile platform, fixed structure, arm body), we calculate $\Vmax$ using the method of \cite{SchSahLat2004}. $\Dmin$ is computed using the FCL library \cite{Pan2012} (BSD License). If the pair is composed of a link and its child link, then the collision checking is disabled, because the two bodies either touch each other by design, or cannot be in collision due to joint bounds. 

\subsection{Collision between a cable and the platform}\label{sec:cable-platform}

Because each cable $i$ is by design in contact with the platform at its attachment point, the method of \cite{SchSahLat2004} used for two bodies of the robot cannot be used as it stands. We choose to simply consider a shortened version of the cable that stops at a fixed distance $d$ of the cable attachment point $B_i$ on the platform. We calculate $\Vmax$ an upper bound on the velocity of any point of the cable relative to the platform, and $\Dmin$, a lower bound on the distance between the shortened cable and the platform.

A previous iteration of our algorithm used a sampling-based model, computed off-line, to reduce the computing time, but at the expanse of the accuracy.

\textbf{Calculation of a velocity upper bound}

We consider a \straightpath{}, for which the linear and angular velocities of the platform are constant and known as stated in Section \ref{continuous-collision-checking}. They are respectively noted $\VP$ and $\WP$, and we note $\wP = \norm{\WP}$ and $\vP = \norm{\VP}$. $C$ is the origin of the platform frame. We want to compute an upper bound on the velocities of all the points of cable $i$ in the platform frame.

For cable $i$, we consider an orthogonal local frame $\mathscr{F}_i^B$ centered on $B_i$ with its x-axis aligned with $\overrightarrow{B_i A_i}$ and directed toward $A_i$. Let $P_i$ be the point on cable $i$ at a fixed distance $r$ of $B_i$. The relation between the coordinate vector $P_i^p$ of $P_i$ in the platform frame and its coordinate vector $P_i^i$ in the local cable frame $\Fi$ is given by (\ref{eq:3}).

\vspace{-2mm}
\begin{equation} \label{eq:3}
\left( \begin{array}{l}
P_i^p \\
1
\end{array} \right)
 = M_{i/p} 
\left( \begin{array}{l}
P_i^i \\
1
\end{array} \right)
\end{equation}

$M_{i/p} = \left( \begin{array}{cc}
R_{i/p} & T_{i/p} \\
0 \ 0 \ 0 & 1
\end{array}
\right) $ is the homogeneous matrix representing the position and orientation of the local cable frame in the platform frame. $R_{i/p} \in SO(3)$ is a rotation matrix. $T_{i/p} = \overrightarrow{C B_i} = \Bi \in \mathbb{R}^3$ is the position of the attachment point $B_i$ in the platform frame. Since $B_i$ is fixed in the platform frame, $T_{i/p}$ is constant. With $P_i^i = (r \ 0 \ 0)^T$ being constant, by differentiating  (\ref{eq:3}) with respect to time, we get

\vspace{-2mm}
\begin{equation}\label{eq:4}
\left( \begin{array}{c} \dot{P_i^p} \\ 0 \end{array} \right) =
\left( \begin{array}{cc}
[ \Wip ]_{\times} R_{i/p} & 0 \\
0 \ 0 \ 0 & 0
\end{array} \right)
\left( \begin{array}{c}
P_i^i \\
1
\end{array} \right)
\end{equation}

where $\dot{P}_i^p$ is the velocity of point $P_i$ in the platform frame, $[ \Wip ]_{\times}$ is the antisymmetric matrix corresponding to the cross product with the cable frame angular velocity vector $\Wip \in \mathbb{R}^3$ in the platform frame, of norm $\wip$. This gives:

\vspace{-2mm}
\begin{equation}
\dot{P_i^p} = [ \Wip ]_{\times} R_{i/p} P_i^i
\end{equation}

Therefore, we can bound the norm:

\vspace{-2mm}
\begin{equation}\label{eq:pip}
|| \dot{P_i^p} ||_2 \leq \wip || P_i^i ||_2
\end{equation}

By composition of angular velocities, we have $\Wip = \Wio + \Wop$, and by bounding the norm: $\wip \leq \wio + \wP$, with $\W_{i/0}$ being the angular velocity of the cable joint frame $i$ in the global reference frame.

Since $B_i$ is fixed in the platform frame, which has known fixed linear and angular velocities $\VP$ and $\WP$, $B_i$ is moving around $A_i$ with a linear velocity of norm $\vBi \leq \vP + \wP \bi$. Since $ \| \overrightarrow{A_i B_i} \|_2 \geq \Lmin $, we get (\ref{eq:wio}) and then (\ref{eq:wip}).

\begin{equation}\label{eq:wio}
\w_{i/0} \leq \frac{v_p + \omega_p b_i}{\Lmin}
\end{equation}

\begin{equation}\label{eq:wip}
\wip \leq \wP + \frac{v_p + \omega_p b_i}{\Lmin}
\end{equation}

From (\ref{eq:pip}) and (\ref{eq:wip}), and knowing we also have an upper bound on $\norm{P_i^i}$ which is $\Lmax$, we get:

\begin{equation}\label{eq:vmax}
|| \dot{P_i^p} ||_2 \leq \Limax \left(
\frac{\vP + \wP \bi}{\Limin}
+ \wP
\right) = \Vmax
\end{equation}

where $b_i$ is a known constant value depending on the design of the CPDR, and $\Lmin$ and $\Lmax$ can be considered either as constant values depending on the shape of the workspace, or calculated for each \straightpath{}. Eq. (\ref{eq:vmax}) gives an upper bound on the velocity of point $P_i$ in the platform reference frame along a \straightpath{}.


\textbf{Calculation of a distance lower bound}

The FCL library is used to compute a lower bound $\Dmin$ on the distance between a cable $i$ and the mobile platform. A STL file represents the 3D collision model of the platform, while the cable is represented by a cylinder. Since the cable and the platform are connected at point $B_i$, as stated above, we consider a shortened portion of the cylinder $A_i \tilde{B}_i$ where $B_i \tilde{B_i} = d \frac{\overrightarrow{B_i A_i}}{|| \overrightarrow{B_i A_i} ||}$. $d$ is a fixed distance chosen so that for any configuration, if a point of $[B_i \tilde{B}_i]$ is in collision with the platform, then at least a point of $[\tilde{B}_i A_i]$ is also in collision with the platform. By design, $\Dmin$ will never be greater than $d$, and it should be noticed that the closest point between the platform and the cable will generally be the (virtual) end of the cable $\tilde{B}$, i.e. $\Dmin = d$. Choosing $d$ as large as possible reduces the number of iterations necessary to validate an interval, and thus the computing time.

\subsection{Collision between two cables}

If two cables have the same attachment points on the platform or the same exit points on the base structure, collision checking between them is disabled since these two cables cannot collide. For all other cable pairs, a velocity upper bound $\Vmax$ and a distance lower bound $\Dmin$ are calculated and used in (\ref{eq:halflength}).

\textbf{Calculation of a velocity upper bound}

We need to find an upper bound on the velocities of all the points of cable $i$ in the local frame of cable $j$. Let $P_i$ be the point on cable $i$ at a fixed distance $r$ of $B_i$, with $P_i^j$ its coordinate vector in the local frame of cable $j$. Similarly to the calculations in Section \ref{sec:cable-platform}, we can write:
\vspace{-1.5mm}
\begin{equation}
\left( \begin{array}{l}
P_i^j \\
1
\end{array} \right)
 = M_{i/j} 
\left( \begin{array}{l}
P_i^i \\
1
\end{array} \right)
\end{equation}

where $M_{i/j} = \left( \begin{array}{cc}
R_{i/j} & T_{i/j} \\
0 \ 0 \ 0 & 1
\end{array}
\right) $ denotes the homogeneous matrix defining the position and orientation of the local frame of cable $i$ with respect to the local frame of cable $j$. $\W_{i/j}$ is the rotation matrix of norm $\w_{i/j}$ of the local frame of cable $i$ relative to the local frame of cable $j$.

Differentiating with respect to time, and knowing that $P_i^i = (r \ 0 \ 0)^T$ and $T_{i/j} = \overrightarrow{B_j B_i}$ are constant:
\vspace{-1mm}
\begin{equation}
\left( \begin{array}{c} \dot{P}_i^j \\ 0 \end{array} \right) =
\left( \begin{array}{cc}
[ \W_{i/j} ]_{\times} R_{i/j} & 0\\
0 \ 0 \ 0 & 0
\end{array} \right)
\left( \begin{array}{c}
P_i^i \\
0
\end{array} \right)
\end{equation}

We then have $\dot{P}_i^j = [\W_{i/j}]_{\times} R_{i/j} P_i^i$. This gives $\| P_i^j \|_2 \leq \w_{i/j} \| P_i^i \|_2$. We have $\W_{i/j} = \W_{i/0} - \W_{j/0}$. Using (\ref{eq:wio}) and knowing we also have an upper bound on $\norm{P_i^i}$ which is $\Lmax$,, we obtain:

\vspace{-3mm}
\begin{equation}\label{eq:cablecableupperbound}
\| \dot{P}_i^j \|_2 \leq \Limax \left( \frac{\vP + \wP \bi}{\Limin} + \frac{\vP + \wP b_j}{\Ljmin}  \right) = \Vmax
\end{equation}

\textbf{Calculation of a distance lower bound}

We use the FCL library to compute a lower bound on the distance between two cables, represented by cylinders. The radius of the cylinder depend on the actual diameter of the cable, and can take into account a safety margin if desired.
For two cables, the distance between their attachment points on the platform is a fixed upper bound on the distance between the cables and thus on $\Dmin$. It means that the proposed method is slower for a CDPR with cables attached close to one another on the platform than for a CDPR with cables attached far from one another. Indeed, if the cable attachment points are close, $\Dmin$ is small and the algorithm can only validate small portions of the path.

\subsection{Collision between a cable and the robotic arm}

We consider a cable $i$ and a body of the robotic arm, attached to joint $J_a$. Let $J_0 = J_p$, $J_1$, ..., $J_{m-1} = J_a$ the list of joints linking the platform joint $J_p$ to $J_a$. $J_1$ is the joint attaching the robotic arm to the platform.

\textbf{Calculation of a velocity upper bound}

Let $P_i$ be a point on cable $i$, and $P_i^a$ its coordinate vector in the joint frame $J_a$.

\vspace{-3mm}
\begin{equation}\label{eq:cable-solid-1}
\left( \begin{array}{l}
P_i^a \\
1
\end{array} \right)
= M_{j_{m-2}/j_{m-1}} M_{j_{m-3}/j_{m-2}} \ ... \ M_{j_0/j_1} M_{i/p}
\left( \begin{array}{l}
P_i^i \\
1
\end{array} \right)
\end{equation}

Where $M_{j_k / j_{k+1}} = \left( \begin{array}{cc}
R_{j_k / j_{k+1}} & T_{j_k / j_{k+1}} \\
0 \ 0 \ 0 & 1
\end{array} \right)$ is the homogeneous matrix representing the position of joint $J_k$ in the reference frame of $J_{k+1}$.

By differentiating (\ref{eq:cable-solid-1}) with respect to time, we obtain

\vspace{-3mm}
\begin{equation}\label{eq:ineq-vel1}\resizebox{1\hsize}{!}{$
\begin{split}
	\left(\begin{array}{c}
	\dot{P}_i^a \\
	0
\end{array}\right)
& =
\left(\begin{array}{cc}
	\crossprod{\omega_{j_{m-2}/j_{m-1}}} R_{j_{m-2}/j_{m-1}} & \V_{j_{m-2}/j_{m-1}} \\
	0 \ 0 \ 0 & 0
\end{array}\right) \cdots \\
&  \qquad \qquad \qquad \qquad \qquad \qquad \qquad \cdots M_{j_0/j_1} M_{i/p}
\left(\begin{array}{c}
	P_i^i \\
	1
\end{array}\right) \\
& + M_{j_{m-2}/j_{m-1}}\left(\begin{array}{cc}\crossprod{\omega_{j_{m-3}/j_{m-2}}} R_{j_{m-3}/j_{m-2}} & \V_{j_{m-3}/j_{m-2}} \\ 0 \ 0 \ 0 & 0\end{array}\right) \cdots \\
& \qquad \qquad \qquad \qquad \qquad \qquad \qquad \cdots M_{j_0/j_1}  M_{i/p}\left(\begin{array}{c}P_i^i\\1\end{array}\right) \\
& + \cdots \\
& + M_{j_{m-2}/j_{m-1}} \cdots M_{j_0/j1}\left(\begin{array}{cc}\crossprod{\omega_{i/p}} R_{i/p} & \V_{i/p} \\ 0 \ 0 \ 0 & 0\end{array}\right)\left(\begin{array}{c} P_i^i\\1\end{array}\right)
\end{split}$}
\end{equation}

Then by bounding the norm using properties of rigid-body transformations:
\begin{equation}\resizebox{1\hsize}{!}{$
\begin{split}
|| \dot{P}_i^a ||_2 & \leq \ v_{i/p} + \w_{i/p} || P_i^i ||_2 \\
 & + v_{j_0/j_1} ( ||P_i^i||_2 + ||T_{i/p}||_2) \\
 & + v_{j_1/j_2} ( ||P_i^i||_2 + ||T_{i/p}||_2 + ||T_{j_0/j_1}||_2) \\
 & \cdots \\
 & + v_{j_{m-2}/j_{m-1}} ( ||P_i^i||_2 + ||T_{i/p}||_2 + . . . + ||T_{j_{m-3}/j_{m-2}}||_2) \\
\end{split}$}
\end{equation}

We note $D_k$ the cumulative length of joint $J_k$:

\vspace{-5mm}
\begin{equation}
D_0 = 0 \ \ \ \ D_k = \sum_{t=0}^{k-1}|| T_{j_t/j_{t+1}} ||_2 \ \ \text{for} \ k \geq 1
\end{equation}

Since $\Limax$ being an upper bound on $|| P_i^i ||_2$, the origin $B_i$ of the local cable frame is fixed in the platform frame so $v_{i/p} = 0$, and using (\ref{eq:wip}), we obtain an upper bound on the velocity of all points of cable $i$ relative to the frame of joint $J_a$:

\vspace{-6mm}
\begin{equation}
\begin{split}
|| \dot{P}_i^a ||_2 \leq \ \Vmax & = \left( \wP + \frac{v_p + \wP b_i}{\Lmin} \right) \Limax + \\
& \sum_{k=0}^{m-2} \left(
v_{j_k/j_{k+1}} ( \Limax + ||T_{i/p}||_2 + D_k)
\right)
\end{split}
\end{equation}
\vspace{-3mm}

$||T_{i/p}||_2$ and all the $||T_{j_k/j_{k+1}}||_2$ are known by design of the robot. The same reasoning can be applied to validate collisions between a cable and an object of the environment.

\textbf{Calculation of a distance lower bound}

The FCL library is used once again to compute a lower bound on the distance between a cable, represented as a cylinder, and the body of the robotic arm, represented by a 3D collision model.

%% file: experimental-results.tex
We have implemented the continuous collision checking method proposed in Sections \ref{continuous-collision-checking} and \ref{boundcalculation} in the software HPP. The CoGiRo CDPR is simulated in HPP. The continuous method is compared to a discretized collision checking method, which uses a configuration validation method and a time step $\tau$.

The computer used to run the software HPP is an "Intel(R) Core(TM) i7-7600U CPU @ 2.80GHz" with 4,096 KB of cache memory and 16 GB of RAM.


This benchmark is performed on CoGiRo with the robotic arm. Random \straightpaths{} are generated by shooting random configurations in the configuration space, and keeping the first two valid configurations. Each \straightpath{} is validated using the continuous method and the discretized method with different time steps. Results are put into five categories:
\begin{itemize}
\item True positive: a collision was both detected by the discretized and the continuous methods.
\item True negative: no collision were detected either by the discretized nor the continuous method.
\item New true positive: a real collision was found by the continuous method, but was not detected by the discretized method.
\item False positive: a collision was detected by the continuous method but is not an actual collision as determined by the configuration validation method.
\item False negative: the continuous method failed to detect a collision that was found by the discretized method.
\end{itemize}

Results are resumed in Tables \ref{table:random_paths} and \ref{table:computing_times} with different time steps tested for the discretized method. The smaller the time step is, the bigger the average computation time is for the discretized method.

As shown in Table \ref{table:random_paths}, there is no false positive, thanks to the fact that the FCL library only returns zero as a lower bound of the distance between two objects if they are actually in collision. The continuous method is garanteed to find every collision, thus there is no false negative. The discretized method misses collisions that the continous method is able to find, even when the time step is reduced. With a time step of $0.001$s, the discretized method has a longer computing time and fails to detect collisions. These results show the efficiency of the continuous method, and its usefulness in situations where no collision is tolerated. 

\begin{center}

\begin{table}
\resizebox{\columnwidth}{!}{%
  \begin{tabular}{|l||c|c|c|c|c|}
    \hline
    $\tau$ (s) & True pos & True neg & New true pos & False pos & False neg\\
    \hline
    & & & & & \\[-1em]
    0.1 & 633 & 362 & 5 & 0 & 0 \\
    \hline
    & & & & & \\[-1em]
    0.01 & 634 & 362 & 4 & 0 & 0\\
    \hline
    & & & & &\\[-1em]
    0.001 & 634 & 362 & 4 & 0 & 0\\
    \hline
  \end{tabular}%
}
  \\
  \caption{Results for the continuous validation compared to the discretized validation with a time step $\tau$ for 1000 random \straightpaths{}}
  \label{table:random_paths}
\vspace{-8mm}
\end{table}

\end{center}